\documentclass{article}
\usepackage{arxiv}
\usepackage{url}
\usepackage{graphicx}

\usepackage{tikz}
\usepackage{amsmath}

\usepackage[linesnumbered,ruled,vlined]{algorithm2e}
\usepackage{enumitem} 
\usepackage{color}
\usepackage{booktabs}
\usepackage{multirow}
\usepackage{array}
\usepackage{graphicx}
\usepackage{epstopdf}
\usepackage{float}
\usepackage{svg} 
\usepackage{amsmath}
\usepackage{fancyhdr}
\usepackage{colortbl}

\usepackage{algorithmicx}
\usepackage{algpseudocode}
\usepackage[justification=raggedright]{caption}

\definecolor{QTM}{RGB}{255,241,204}
\definecolor{SLM}{RGB}{216,234,210}
\definecolor{PCM}{RGB}{247,225,237}

\begin{document}

\title{Can't say cant? Measuring and Reasoning of Dark Jargons in Large Language Models}
%

\author{{\hspace{1mm}Xu Ji, {\includegraphics[scale=0.06]{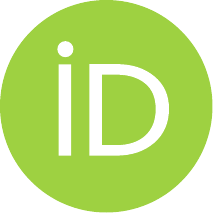}\hspace{1mm}Jianyi Zhang}\thanks{Corresponding author: \texttt{zjy@besti.edu.cn}},  Ziyin Zhou, Zhangchi Zhao, Qianqian Qiao} \\
	Beijing Electronic Science and Technology Institute\\
	Beijing, China 100070 \\
	\And
	{\hspace{1mm}Kaiying Han} \\
	Xinan Jiaotong University\\
	Chengdu, China \\
 \And
	{\hspace{1mm}Md Imran Hossen, Xiali Hei} \\
	University of Louisiana at Lafayette\\
	Louisiana US 70503 \\
}

\maketitle              

\begin{abstract}
Ensuring the resilience of Large Language Models (LLMs) against malicious exploitation is paramount, with recent focus on mitigating offensive responses. Yet, the understanding of cant or dark jargon remains unexplored. This paper introduces a domain-specific Cant dataset and CantCounter evaluation framework, employing Fine-Tuning, Co-Tuning, Data-Diffusion, and Data-Analysis stages. Experiments reveal LLMs, including ChatGPT, are susceptible to cant bypassing filters, with varying recognition accuracy influenced by question types, setups, and prompt clues. Updated models exhibit higher acceptance rates for cant queries. Moreover, LLM reactions differ across domains, e.g., reluctance to engage in racism versus LGBT topics. These findings underscore LLMs' understanding of cant and reflect training data characteristics and vendor approaches to sensitive topics. Additionally, we assess LLMs' ability to demonstrate reasoning capabilities. Access to our datasets and code is available at https://github.com/cistineup/CantCounter.
\end{abstract}
\section{Introduction}
Large Language Models (LLMs), exemplified by ChatGPT\cite{TP-OpenAI}, redefine information acquisition, communication, and problem-solving\cite{ray2023chatgpt}. These models are trained on extensive datasets or fine-tuned from pre-existing models, necessitating vast amounts of data. However, LLMs also pose security and ethical concerns as attackers can exploit their generative capabilities for malicious purposes \cite{roy2023generating}. Such misuse encompasses disinformation dissemination \cite{weidinger2022taxonomy}, AI-driven crime \cite{birks2023linking}, privacy breaches \cite{li2023multi}, and social engineering \cite{gupta2023chatgpt}. Despite efforts by regulators like OpenAI to implement content filters \cite{TP-OpenAI-classifiers}, there remains a risk of attackers disguising malicious content using ``cant" or ``dark jargon" - concealed language elements requiring deeper comprehension \cite{yuan2018reading}. LLMs excel in understanding and generating natural language responses, fostering user trust. While research evaluates their efficacy in providing accurate responses \cite{tan2023can}, little attention has been paid to LLMs' interaction with cant in specific domains. Prior studies often lack depth in understanding the intricacies of cant \cite{rozado2023political}, especially its varied representations in domains like politics and drugs. In this paper, we investigate LLMs' ability to recognize and reason about cant, particularly in domains prone to offensive content like politics and drugs. Despite progress in filtering harmful content, attackers can still exploit cant to evade detection. Understanding LLMs' response to cant in specific domains is essential for addressing emerging security challenges. Additionally, we assess LLMs' ability to demonstrate reasoning capabilities.

\textbf{Research Questions.} To address the above issues, in this paper, we evaluate the reasoning abilities of current LLMs involving cant or dark jargon from the following four perspectives:
\vspace{-2mm}
\begin{itemize}
\setlength{\leftskip}{0.5em}
\item[1.] \textbf{RQ1:} Do different types of questions help LLM understand the cant? 
\item[2.] \textbf{RQ2:} Do different question setups and prompt clues help LLM understand cant?
\item[3.] \textbf{RQ3:} Do different LLMs have the same understanding of the same cant?
\item[4.] \textbf{RQ4:} How well does LLM understand cant in different domains?
\end{itemize}    
\vspace{-2mm}

\textbf{CantCounter:} Addressing past shortcomings\cite{rozado2023political}, CantCounter is a system crafted to evaluate LLM's grasp of cant within specific domains. We compile Cant and Scene datasets from various sources to form adversarial texts. These datasets fine-tune the GPT-2 model and generate Scene fragments for assessing LLM comprehension. Co-Tuning methods align the Cant dataset and Scene fragments, while Data-Diffusion techniques augment and refine adversarial text. Employing Type, Sample learning, and Clue approaches enrich our experiments. Finally, Data-Analysis methods systematically evaluate 1.67 million data points. CantCounter is locally deployable and adaptable to any open-world dialogue system. Its replication has both advantages and drawbacks, aiding attackers in bypassing LLM classifiers while facilitating safety filter development. We define ``\textbf{entities}" as distinct objects or concepts and ``\textbf{scenes}" as related events in specific environments.

\textbf{Ethical Considerations:} CantCounter draws from public datasets such as Reddit \cite{TP-Reddit} and 4chan \cite{TP-4chan}, avoiding direct user interaction. However, its misuse poses risks, despite its benefits in addressing LLM's challenges. Despite these potential risks, we believe that the benefits of CantCounter far outweigh the risks. LLM has become a hot topic \cite{sohail2023decoding}, and we need to fully recognize the potential problems of LLM and promote its safer development and application. We caution that \textcolor{red}{\textbf{this paper may contain sensitive content, including drug and violence-related examples, which could cause discomfort.}} Comprehensive data is available upon request. We have only open sourced part of the dataset.

\textbf{Contributions.} This paper introduces three key contributions:
\vspace{-2mm}
\begin{enumerate}
\item We present the Cant and Scene datasets, addressing data scarcity in domains like drugs, weapons, and racism, laying groundwork for future large language model assessment.
\item CantCounter, our framework, assesses large language models' understanding of domain-specific cants through four stages: Fine-Tuning for scene fragment generation, Co-Tuning for cross-matching, Data-Diffusion for text expansion, and Data-Analysis for simplifying complex calculations.
\item Our evaluation of CantCounter reveals its efficacy in bypassing security filters of mainstream dialogue LLMs, providing insights into LLM reasoning within specific domains and guiding future research. 
\end{enumerate}

\section{Background}\vspace{-2mm}
\subsection{Large Language Model Security Issues}
\vspace{-2mm}
ChatGPT, developed by OpenAI in November 2022 \cite{TP-OpenAI}, has undergone upgrades and fine-tuning \cite{RLHF} to prevent harmful content generation. However, users can still provoke negative responses by using specific prompts \cite{wang2023adversarial}. Researchers are investigating security risks, including the generation of toxic outputs from benign inputs \cite{zhangyang-Why-so-toxic}. Recent studies have shown that attackers can bypass detection by encrypting inputs with methods like Caesar ciphers and exploiting language nuances \cite{yuan2023gpt}. This paper proposes a Q\&A query approach to evaluate LLMs' reasoning abilities in handling such content.

\vspace{-3mm}
\subsection{Cant}
\vspace{-2mm}
Cant, a specialized language used by social groups for secrecy \cite{ZFXK202103013}, varies in names like argot \cite{sourdot1991argot}, slang \cite{wu2018slangsd}, and secret language across history. While LLMs excel in traditional cant analysis, understanding criminal cant poses challenges. Criminal groups use innocuous terms to hide illegal activities, necessitating mastery for law enforcement \cite{WHXU201402006}. Our study explores cant in politics, drugs, racism, weapons, and LGBT issues. These cants share ambiguity, indirect messaging, and potential for social harm. Political cant conveys biases, drug cant evades regulation, racism cant reinforces biases, weapons cant enables illegal dealings, and LGBT cant discriminates. Mastering these cants is vital for addressing societal and security concerns.
\vspace{-3mm}
\subsection{Question Answering (Q\&A) Task}
\vspace{-2mm}
Dialogue systems fall into task-oriented and non-task-oriented categories. Task-oriented systems serve specific purposes like reservations, while non-task-oriented systems engage in free conversation. Examples include ChatGPT, Bard, ERNIE, and Claude, offering services in entertainment, social interaction, and information retrieval \cite{yan2017building}.Question-answering (Q\&A) tasks in NLP evaluate language processing capabilities \cite{dasigi2021dataset}, including reading comprehension and logical reasoning. Q\&A formats include abstractive, Yes/No, and Multiple-Choice, each requiring specific evaluation metrics \cite{rogers2023qa}. We employ Zero-shot/One-shot learning for testing.

\section{CantCounter}
\vspace{-2mm}
\subsection{High-level Idea}
\vspace{-2mm}
We observe that the responses generated by LLMs vary with different cants, allowing adversaries to bypass filters or security restrictions. Thus, understanding how LLMs react to different cants is very important. However, exhaustively trying different cants queries with different scenes across numerous domains to find those capable of bypassing LLM restrictions and generating harmful outputs would be time-consuming and impractical. Therefore, we investigate whether adversaries can independently combine different cants and scenes to generate context that is reasonable and coherent, bypassing LLM filters or restrictions. To this end, we introduce CantCounter, the first evaluation (attack) framework targeting open-world dialogue systems (LLM).
\vspace{-3mm}
\subsection{Threat Model} \label{sec:4a}\vspace{-2mm}
We adopt a threat model similar to ``Why so toxic" \cite{zhangyang-Why-so-toxic}, targeting deployed dialogue LLMs like ChatGPT. Firstly, the adversary requires scene data different from the target LLM's training data. Secondly, they interact with the LLM, combining cants and scenarios to extract detectable cants. Finally, they access the victim LLM via CantCounter in a black-box manner, querying it through an API-like interface.

\vspace{-3mm}
\subsection{Dataset}\vspace{-2mm}
In our study, we extensively gathered cant related to five domains: politics, drugs, racism, weapons, and LGBT. The cant, comprising common and less common usages, holds practical meanings in real life. This Cant dataset forms a robust basis for evaluating the veracity and reliability of LLMs across specific domains. These five areas were chosen to address pressing societal issues impacting fundamental values such as social justice and human rights. Exploration of politics, drugs, racism, weapons, and homosexuality enables LLMs to tackle real-world challenges effectively. While other domains like hacking and fraud are significant, we focused on these due to data availability and processing feasibility, leaving room for future research on sensitive topics.

\begin{figure*}[htbp]
\centering
\includegraphics[scale=0.3]{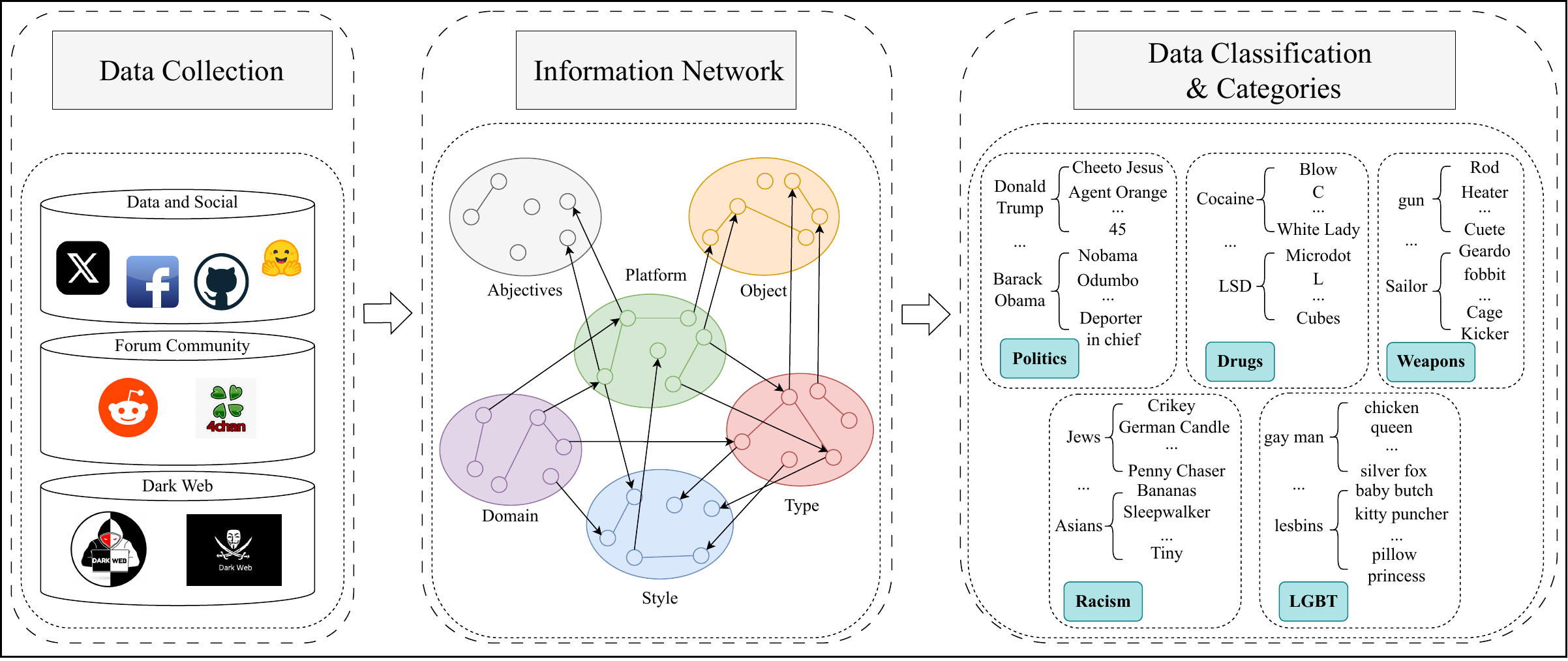} \vspace{-2mm}
\caption{Construction of the Cant dataset involves collecting, summarizing security-related data, forming interconnected relationships between cants into an information network, and establishing the dataset through data classification and categorization, encompassing various domain-related entities and their corresponding cants.}
\label{fig2} \vspace{-3mm}
\end{figure*}

In constructing the Cant dataset (Figure \ref{fig1} \normalsize{\textcircled{\scriptsize{2}}}\normalsize), we crawled or manually screened multiple sources, including government agency websites \cite{TP-Eric-Patterson}, online forums like Reddit \cite{TP-Reddit}, 4chan \cite{TP-4chan}, and  X \cite{TP-X-Corp}, publicly available datasets from Kaggle \cite{TP-Kaggle} and Hugging Face \cite{TP-Hugging-Face}, dark web, and public compilations of cant. Multi-source data encompasses various text types closely related to specific domains. CantCounter utilizes information networks \cite{zhao2020cyber} to address redundancy challenges between cants, capturing their interdependency.

The Cant dataset covers five domains, totaling 1,778 cants across 187 entities. We randomly selected 53 entities, totaling 692 cants, ensuring even representation across domains and prevalence in the open world. Selected entities and cants were cross-validated with authoritative sources \cite{TP-politics1,TP-drug1,TP-Weapon2,TP-race1,TP-LGBT1} to ensure wide presence and reflection in publicly accessible information sources. Criteria like content relevance and topic specificity guided information selection and filtering, aiming for transparency and consistency. The resulting high-quality data forms the Scene dataset, laying the groundwork for subsequent simulation scene generation models.

During information selection and filtering (Figure \ref{fig1} \normalsize{\textcircled{\scriptsize{1}}}\normalsize), explicit criteria were used to judge relevance and adherence to study definitions. Decisions were reached through participatory discussion to mitigate subjectivity and ensure alignment with research objectives. This rigorous process yields a refined dataset for accurate and relevant analysis.

\vspace{-3mm}
\subsection{Pipeline}
\vspace{-2mm}

\begin{figure}[htbp]
\centering
\includegraphics[scale=0.35]{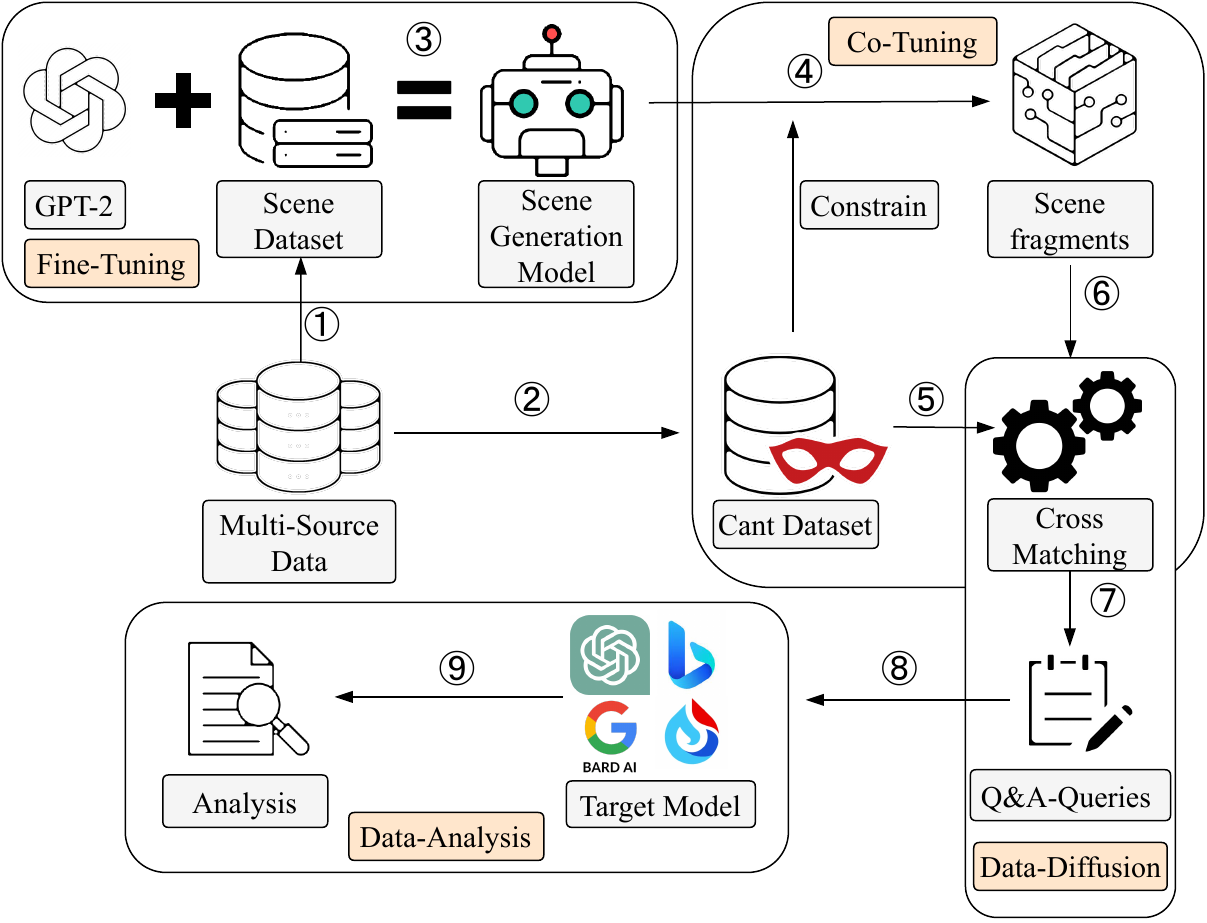}\vspace{-2.5mm}
\caption{The pipeline of CantCounter.}
\label{fig1} \vspace{-2mm}
\end{figure}

The CantCounter pipeline (Figure \ref{fig1}) consists of four stages: \textbf{Fine-Tuning}, \textbf{Co-Tuning}, \textbf{Data-Diffusion}, and \textbf{Data-Analysis}, as detailed below.

Cant is prevalent in the open world, so we aggregate raw text data from various sources to construct Cant and Scene datasets (Section 3.3). Although Cant and Scene datasets provide specific entities and scenes, they may not align well with the domain's requirements. Therefore, in Stage \textcircled{\scriptsize{3}}, we fine-tune GPT-2 using the Scene dataset to build five scene generation models for large-scale scenes, tailored to our specific domains. However, the fine-tuned scenes may not match the entities in the Cant dataset. In Stage \textcircled{\scriptsize{4}}, we address this issue by using entities from the Cant dataset to constrain the output of the generated model, ensuring scenes closely relate to the cant entities. Next, we conduct semi-automatic screening of the generated simulation scenes to form a set of Scene fragments. While these fragments contain entities, linking them with specific questions requires a method we have not yet discovered. Hence, in Steps \textcircled{\scriptsize{5}}-\textcircled{\scriptsize{6}}, we devise the Co-Tuning stage, where Scene fragments cross-match with cants from the Cant dataset to form Fragments. To enable multi-task comparison, we construct detection tests through different combinations of specific domains, question types, learning methods, and prompt clue methods in Stage \textcircled{\scriptsize{7}}. This completes and diffuses Fragments to form Q\&A-Query datasets.

Finally, in Stages \textcircled{\scriptsize{8}}-\textcircled{\scriptsize{9}}, Q\&A-Queries are sent to the target model API for completion, and a segmented data statistics algorithm is applied to obtain and analyze test results, conducting analyses in the Data-Analysis stage.

\vspace{-3mm}
\subsection{Stage 1: Fine-Tuning}\vspace{-2mm}

During the fine-tuning stage, we use the Scene dataset to guide GPT-2 in generating tailored scenarios for specific domains. Despite more advanced models like GPT-3.5 and GPT-4 being available, we opt for GPT-2 due to its open-source nature, facilitating better control over training details. The fine-tuning code is publicly accessible for replication. The fine-tuning process is outlined in Algorithm \ref{alg1}.

The Transformer model \cite{vaswani2017attention} forms the basis for GPT-2, featuring encoders and decoders with identical modules. GPT-2 employs a partially masked self-attention mechanism and positional coding to understand sequence relationships. It has been successfully applied in various tasks like AI detection and text summarization. Overall, GPT-2's fine-tuning with the Scene dataset enables the generation of Question-Answer patterns tailored to specific domains, aiding in simulated scene generation tasks.
\vspace{-3mm}
\begin{algorithm} 
\small  
    \caption{Fine-Tuning} 
    \label{alg1} 
        \KwIn{ pre-trained model parameters $\theta_\mathrm{p}$, fine-tuned dataset $D_f$, loss function $L(\theta)$ that depends on the model parameters $\theta$, optimizer $O(\theta)$ for updating the parameters $\theta$, learning rate $\eta$, number of iterations $T$ for fine-tuning, $x$ is the input sample, $y$ is the label, $\hat{y}$ is the model's predicted value, and $\nabla\theta$ denotes the gradient operation  
        }
        \KwOut{fine-tuned model parameters $\theta_\mathrm{f}$}

        Initialization $\theta_\mathrm{f}$ = $\theta_\mathrm{p}$
        
        \For{each $t$ \textbf{in} range$(1, T+1)$}
        {
            \For{$batch_x$, $batch_y$ \textbf{in} $D_f$}
            {
                $\hat{y}$ = model$(batch_x,\theta_\mathrm{f})$;
                
                loss = $L(\hat{y}, batch_y)$;
                
                gradient = torch.autograd.grad(loss,$\theta_\mathrm{f}$) ;
                
                $\theta_\mathrm{f}= O(\theta_\mathrm{f}, gradient, \eta) $;
                
            }
        }
        \Return{$\theta_\mathrm{f}$}
\end{algorithm}

\vspace{-3mm}

\subsection{Stage 2: Co-Tuning}\vspace{-2mm}

To solve the problem of many intersecting data processes in CantCounter, we use the Cant dataset and Scene fragments to collaborate and design a Co-Tuning method. Co-Tuning realizes the generation and collaboration of cross-matching and solves the problem of detection data insufficiency. The Cant dataset provides detailed entity information for the generated model. The entities could constrain the generative model and make the Scene Fragments more consistent and coherent in the need for a specific domain during the Co-Tuning stage. In the end, we also manually review the results to ensure the relevance of cants to scenes and the distinctiveness of all scenes corresponding to the same cant.

In this paper, we design formulas in the Co-Tuning to mathematically represent this part of the stage. The generation model is specified as $M_p (p\in[1,5])$, and it includes five fine-tuned models, which are denoted as $M_1$, $M_2$, $M_3$, $M_4$, and $M_5$.  

\begin{figure}[t]
\centering
\includegraphics[scale=0.28]{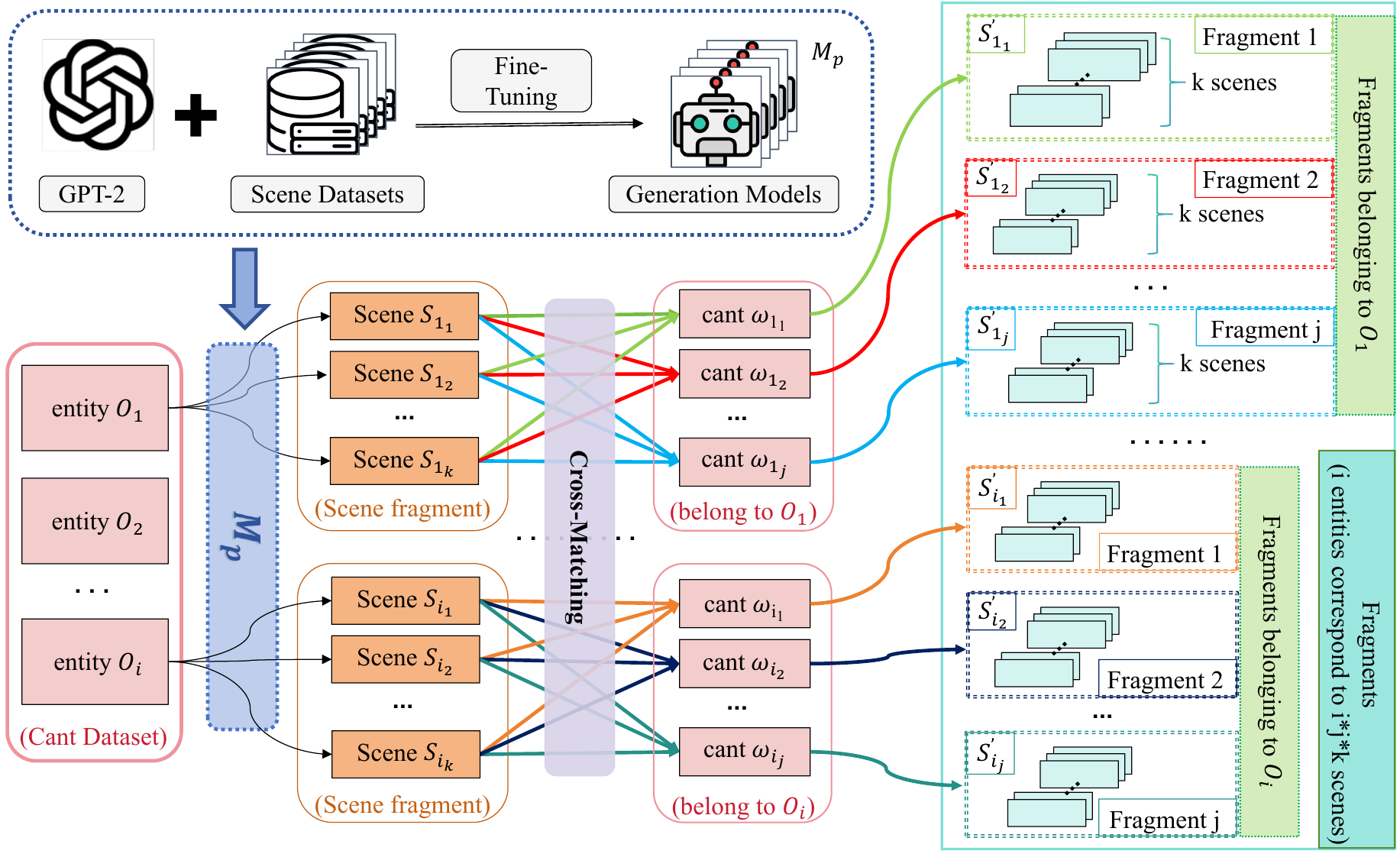}\vspace{-2mm}
\caption{The overall structure and process of Co-Tuning.}
\label{fig3}\vspace{-3mm}
\end{figure}

As shown in Figure \ref{fig3}, entity $O_i$ represents the $i$-th entity ($i \in [1, 15]$) in the Cant dataset, and cant $\omega_\mathrm{i_j}$ represents the $j$-th cant of $O_i$ ($j \in [1, 20]$). For example, in the case of the politics domain, there are 10 entities used in our experiments, each entity has twenty cants, $j$ is taken as $[1,20]$. The entity $O_i$ can constrain the fine-tuned model $M_p$'s output, and the result of the constraint is the Scene fragment; this part corresponds to \textcolor{blue}{Eq. (1)}. The Scene is $S_{i_k}$  $(i\in[1,10],k\in[1,101])$. The Scene $S_{i_k}$ represents the $k$-th scene fragment ($i \in [1, 10]$, $k \in [1, 101]$) that the $i$-th entity enters into the output of the fine-tuning model ($M_p$).

\vspace{-1mm}
\begin{equation}
    S_\mathrm{i_k}=M_p(O_i)
\end{equation}
\vspace{-1mm}

\textcolor{blue}{Eq. (2)} denotes the cross-match of Cant and Scene fragment and was saved in $S_\mathrm{i_k}^{'}$.
\vspace{-1mm}
\begin{equation}
    S_\mathrm{i_k}^{'}=S_\mathrm{i_k}\cup\omega_\mathrm{i_j}
\end{equation}
\vspace{-1mm}
There are $k$ orange boxes in the $O_1$ Scene fragment. These orange boxes represent the $M_p$-generated text containing the Cant dataset's entities. The function of Eq. 2 is to replace the entities in the Scene fragments with cant in the Cant dataset. As shown in Figure \ref{fig3}, for example, from $O_1$ Scene fragment to \textit{Fragment 1}. We replace entities in Scene $[S_{1_1},S_{1_k}]$ with the cant ($\omega_\mathrm{1_1}$), forming \textit{Fragment 1}. By analogy, we built \textit{$j$ Fragments} in the Co-Tuning stage. 

In the Co-Tuning stage, we can obtain scene fragments related to entities in specific domains that have a high degree of context consistency and express various characteristics of the entities in different contexts. At the same time, our fine-tuned model is flexible enough to introduce multiple entities during the generation process and allow scene fragments to describe the relationships among multiple entities. This stage generates diverse scene fragments. While the scene fragments are generated through a generative process, the Scene dataset we provide undergoes manual review to mitigate errors in both the generated content and the language utilized within the experimental environment.

\begin{figure}[htbp]
\centering
\includegraphics[scale=0.3]{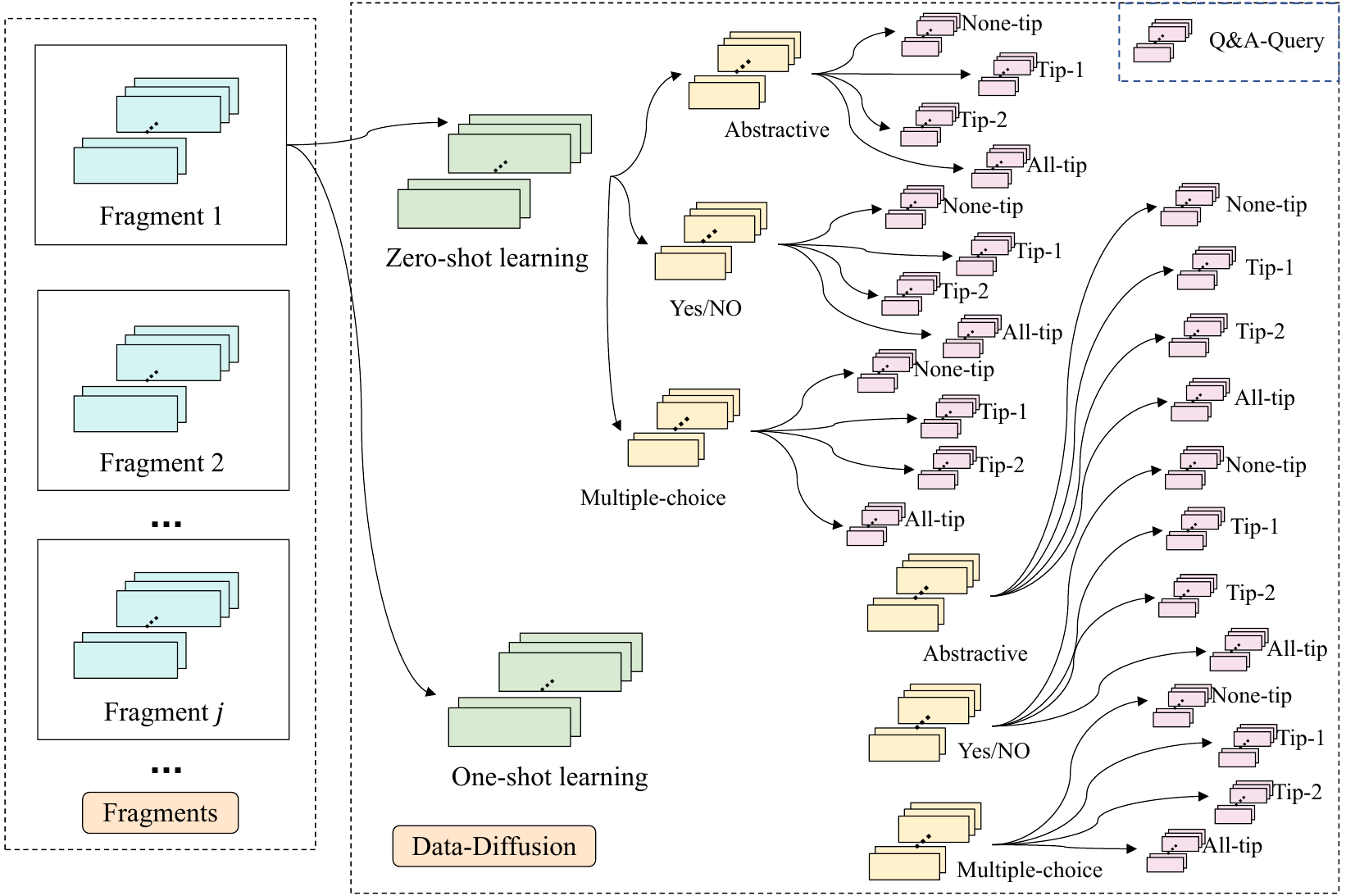}\vspace{-2mm}
\caption{Schematic diagram of Data-Diffusion.}\vspace{-2mm}
\label{fig4}
\end{figure}

\vspace{-3mm}
\subsection{Stage 3: Data-Diffusion}\vspace{-2mm}

At this stage, \textit{Fragments} from the Co-Tuning stage are transformed into Q\&A-Queries to enhance interaction with LLM and diversify evaluation. We employ three diffusion methods: two sample learning techniques, three question types, and four prompt clue methods. Each \textit{Fragment} generates 24 Q\&A-Queries. First, we introduce sample learning techniques for zero-shot and one-shot learning transformations of \textit{Fragments}. Second, we categorize \textit{Fragments} into Abstractive, Yes/No, and Multiple-choice question types. Finally, prompts are classified into None-tip, Tip-1, Tip-2, and All-tip categories, considering information retrieval difficulty and situational prompting.

The introduction of Data-Diffusion in extended \textit{Fragments} has significantly increased Q\&A queries, providing diverse test cases for evaluating the generation model's performance comprehensively. This approach promises to establish a diverse database for future research and applications.

\vspace{-3mm}
\subsection{Stage 4: Data-Analysis}
\vspace{-2mm}
As shown \normalsize{\textcircled{\scriptsize{8}}}\normalsize\ and \normalsize{\textcircled{\scriptsize{9}}}\normalsize\ in Figure \ref{fig1}, \normalsize{\textcircled{\scriptsize{8}}}\normalsize\ means sending the data expanded by Data-Diffusion to ChatGPT and other target models. \normalsize{\textcircled{\scriptsize{9}}}\normalsize\ 
 shows data analysis of the output results of LLMs such as ChatGPT. After completing the Data-Diffusion, we submit the generated Q\&A-Queries to the LLM API interface to obtain a large number of data results. These data results are complex and diverse, including the interplay of relationships. Therefore, we devise a data analysis algorithm to yield both numerical and analysis outcomes.

\begin{algorithm}
\small
\caption{Segmented data statistics algorithm}
\label{alg2}
    \KwIn{\textit{$J$} -- $j$ \textit{Fragments} in Co-Tuning (Fig \ref{fig4}), \text{$K$} -- $k$ Scenes in a \textit{Fragment}, \text{$tasks$} -- [``Abstractive": \{``AZ", ``AO"\}, ``Yes/NO": \{``JZ", ``JO"\}, ``Multiple-choice": \{``MZ", ``MO"\} ], \text{$clues$} --  [``None-tip", ``Tip-1", ``Tip-2", ``All-tip"], \text{$intervals$} --  [0, 1-10, 11-20, ..., 91-101]
    }
    \KwOut{$R_\mathrm{j,t,c}$, $N_\mathrm{j,t,c,z}$, $Sum_\mathrm{t}$, $Sum_\mathrm{PCM}$}

\SetAlgoNlRelativeSize{0}
\SetAlgoNlRelativeSize{-1}

\For{each $j$ \textbf{in} $J$ }{
    \For{each $t$ \textbf{in} tasks }{
        \For{each $c$ \textbf{in} clues}{
          $N_\mathrm{j,t,c}$=$\sum_{a=1}^{k} i_a$
           (\textbf{if} \textit{hit} \textbf{then} $i_a$=1, \textbf{else} $i_a$ = 0)\\           
           $R_\mathrm{j,t,c} \longleftarrow \frac{N_\mathrm{j,t,c}}{k}$
      
            \For{each $z$ \textbf{in} interval}
               {\textbf{if} $N_\mathrm{j,t,c}$ \textbf{in} $z$ \textbf{then}
               $N_\mathrm{j,t,c,z}$++}
            }
        }
    }
    $Sum_\mathrm{t}$=$\sum_{\alpha=1}^{j}\sum_{\beta=1}^{c} N_\mathrm{\alpha,t,\beta}$
  
    $Sum_\mathrm{PCM}$=$\sum_{\alpha=1}^{j}\sum_{\beta=1}^{t} N_\mathrm{\alpha,\beta,c}$
    
\Return $R_\mathrm{j,t,c}$, $N_\mathrm{j,t,c,z}$, $Sum_\mathrm{t}$, $Sum_\mathrm{PCM}$

\end{algorithm}

After the Co-Tuning and Data-Diffusion stages, the test data generated by CantCounter is very complex. Therefore, in the Data-Analysis stage, we implement Algorithm \ref{alg2} to conduct data statistics from various angles. During analysis, when the entity $O_i$ is modified in the Co-Tuning stage (see Figure \ref{fig3}), Algorithm \ref{alg2} will be called accordingly. We analyze the results based on different tasks. We learn and analyze data features from Question Type Method (See 4.2 QTM) and Sample Learning Method (See 4.3 SLM) based on different question types and samples learning to get $Sum_t$; we analyze the data based on different prompt clues from Prompt Clue Method (See 4.4 PCM) to get $Sum_{PCM}$. In Algorithm \ref{alg2}, we set the matching conditions, calculate the number of fragments, and obtain $N_{j,t,c}$ and accuracy $R_{j,t,c}$. At the same time, we set eleven intervals: 0, 1-10, 11-20, ..., 91-101 to distinguish different feedbacks and obtain $N_{j,t,c,z}$.
  
As shown in the Algorithm \ref{alg2}, we put Zero-shot learning, One-shot learning, and three tasks together as a loop. We define that in the Abstractive task, the output is $AZ$ in the Zero-shot learning input; the output is $AO$ in the One-shot learning input. In the Yes/NO task, the output is expressed as $JZ$ in the Zero-shot learning input; the output is expressed as $JO$ in the One-shot learning input. In the Multiple-choice task, the output is represented as $MZ$ in the Zero-shot learning input; the output is expressed as $MO$ in the One-shot learning input. The above content has been integrated into our code to form semi-automation.
\vspace{-4mm}

\section{Experimental Design and Results}\vspace{-2mm}
\label{sec4}

To explore our research questions, we conducted experiments in CantCounter, outlined sequentially in this section. We examined various question types in RQ1 (Section \ref{sec:4b}), different question setups in RQ2 (Section \ref{sec:4c}), and diverse prompt clues in RQ2 (Section \ref{sec:4d}). Focusing primarily on ChatGPT-3.5 (version gpt-3.5-turbo-0613) due to its convenience and wide usage, similar experiments were also conducted with other language models. All experiments were performed on a server equipped with an RTX 3090 Ti GPU. In this section, we analyze using cant and scene to bypass the LLM filter in the CantCounter framework quantitatively. We conduct open-world query experiments across five domains: politics, drugs, racism, weapons, and LGBT. Initially setting $k$ to 101, we match 692 cants to 53 entities, resulting in 69,892 scenes. These undergo Data-Diffusion, expanding to 1,677,408 scenes. This study enables a comprehensive analysis of corpus performance and language changes within specific domains.

\vspace{-3mm}
\subsection{Question Type Method (QTM)} \label{sec:4b}
\vspace{-2mm}

\begin{figure*}[htbp]
\centering
\includegraphics[scale=0.4]{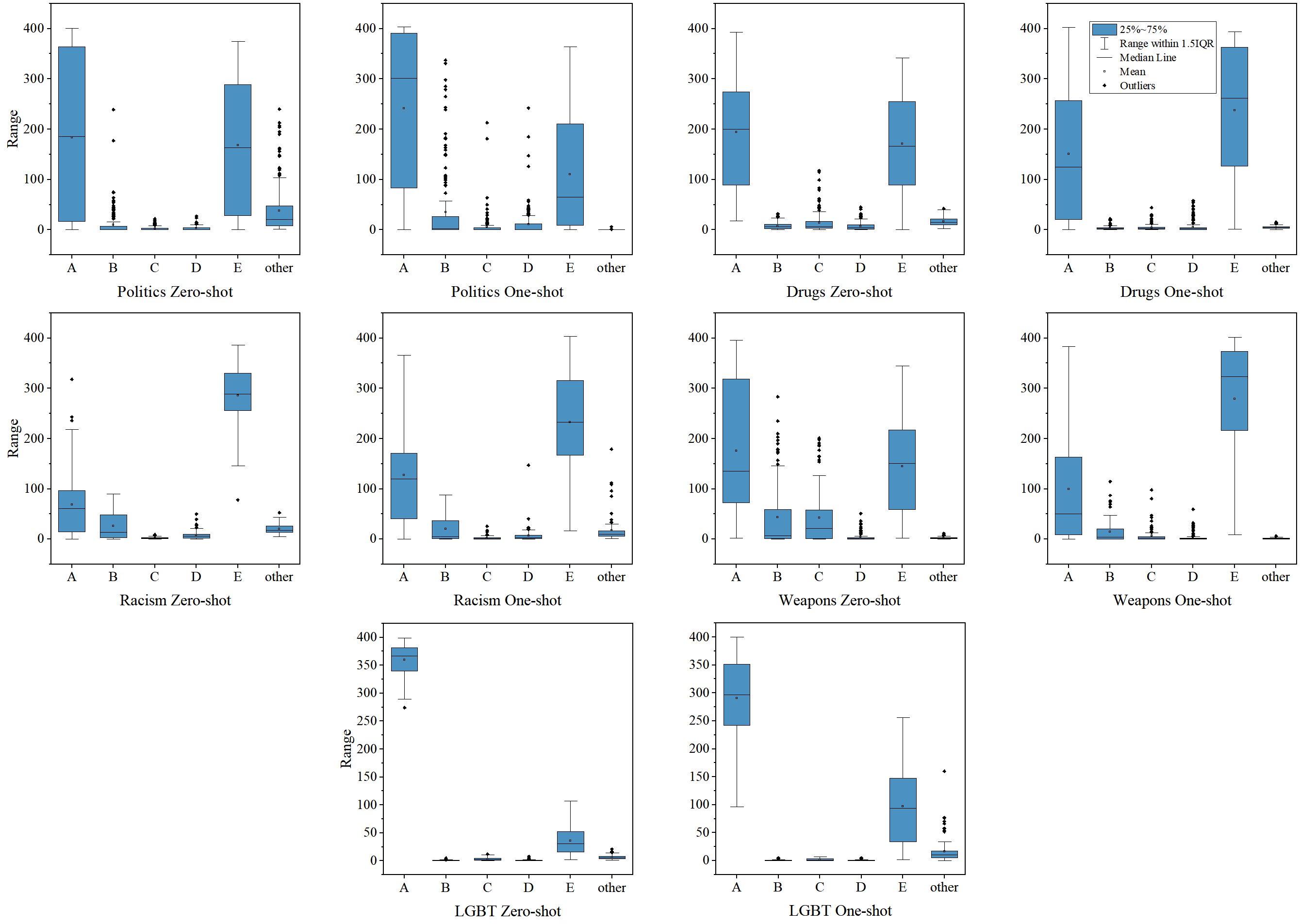}\vspace{-2mm}
\caption{The vertical axis refers to the number of correct answers under the four tips. The total number is 404. (A) and (E) stand out in Multiple-choice, being the correct answer and ``I don't know" respectively. After carefully studying ChatGPT-3.5's interpretation of option (E), we find that when the context is ambiguous or the entities in the implicit context are rare, ChatGPT-3.5's accuracy will drop significantly; that is, it will prefer option (E).}
\label{box}\vspace{-4mm}
\end{figure*}

In the Q\&A task, we conduct three types of tasks:
\vspace{-2mm}
\begin{itemize}
\item \textbf{Abstractive Task}: Models generate responses freely, without relying on specific information extraction.

\item \textbf{Yes/No Task}: Models provide binary responses, ``True" or ``False," based solely on the presented question and existing knowledge.

\item \textbf{Multiple-choice Task}: Models select the correct answer from a set of options, demonstrating comprehension of semantics and accurate identification.
\end{itemize}
\vspace{-2mm}
Table \ref{tab2} shows that Multiple-choice tasks achieve the highest accuracy (45.38\%), while Yes/No tasks have the lowest (22.91\%). The discovery that ChatGPT performs well in multiple-choice questions is intriguing. In this task, there are five options (A) to (E), with (A) to (D) relevant to a specific domain, and (E) set as ``I don't know." ``Other" signifies an answer unrelated to these options, with (A) as the correct choice. Figure \ref{box} displays the box plot analysis results. Analyzing the Multiple-choice task results, we find key factors for its success. Firstly, it offers a set of answers with one correct option and distractors, aiding comprehension. Secondly, its structured format simplifies the process of eliminating incorrect options, improving accuracy. Lastly, the inclusion of an ``I don't know" option enhances accuracy in uncertain situations.

We also explore the low accuracy in the Yes/No task. Comparing ChatGPT-3.5's ``False" answers with Multiple-choice task data, we find they often include option (E) and incorrect choices from the Multiple-choice task due to the clarity of options. Additionally, differences in response styles and keyword detection criteria impact ChatGPT-3.5's performance across Abstractive and Yes/No tasks, where Yes/No tasks restrict responses to ``True" or ``False." Overall, our analysis highlights how different Q\&A types affect ChatGPT-3.5's accuracy in specific domains, with Multiple-choice tasks showing higher performance. Further research is needed to improve ChatGPT-3.5's accuracy and adaptability in these domains.

\vspace{-3mm}
\subsection{Sample Learning Method (SLM)} \label{sec:4c}
\vspace{-2mm}

In our experiments, we explore two sample setups: Zero-shot and One-shot learning.
\vspace{-2mm}
\begin{itemize}
\item \textbf{Zero-shot learning.} No examples are provided in the prompt, which only includes instructions and questions.

\item \textbf{One-shot learning.} The prompt includes an example relevant to the discussion, consisting of a sample message and user information.
\end{itemize}
\vspace{-2mm}
Zero-shot learning involves a single user message, while One-shot learning processes a sample message and a user message. These methods help understand LLM's performance in different sample learning approaches and reveal its inference capabilities in information-poor settings. Further investigation uncovers learning patterns and effects of the model in specific domains, with default hyper-parameter settings used to avoid extensive tuning.

\begin{figure}[H]
\centering
\includegraphics[scale=0.5]{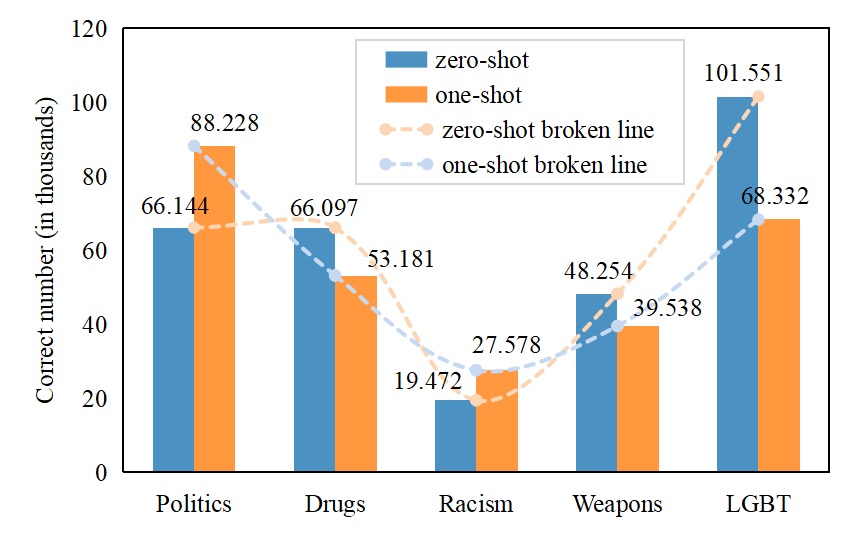}
\caption{Comparison of the number of correct Zero-shot learning and One-shot learning in different domains.}
\label{Zero-vs-One}\vspace{-3mm}
\end{figure}

In this section, we explore how Zero-shot and One-shot learning methods affect LLM accuracy in recognizing cant scenes for RQ2. Traditionally, One-shot learning often outperforms Zero-shot learning due to more available data \cite{zhong2023study}. However, our cross-domain analysis, depicted in Figure \ref{Zero-vs-One} and reflected in Table \ref{tab2} (red section), reveals a trend favoring Zero-shot learning overall. We find this trend varies by domain.

In the politics domain, One-shot learning performs better due to ample data and contextual understanding. Conversely, in the LGBT domain, Zero-shot learning outperforms One-shot learning due to limited publicly available examples. One-shot learning aids ChatGPT-3.5 in better contextual comprehension of sensitive topics, but it may also introduce biases, leading to lower overall accuracy in specific domains. Similar analyses across other domains yield consistent results.

\vspace{-3mm}

\subsection{Prompt Clue Method (PCM)} \label{sec:4d}
\vspace{-2mm}
In this part of the study, the purpose of CantCounter is to explore the impact of different clues on LLM recognition and reasoning abilities. To this end, we provide four different clues to experiment with:
\begin{itemize}
\vspace{-2mm}
\item \textbf{None-tip.} Keeps the same as the original prompt and does not add any additional clues.

\item \textbf{Tip 1.} Add relevant tip for ``None-tip". For example, when describing Trump's cant, we can add the clue ``Politician" in the political domain to make the prompt more directional.

\item \textbf{Tip 2.} Add another relevant tip for ``None-tip". For example, when describing Trump's cant, add the ``United States" prompt in the domain of politics to enrich the prompt content.

\item \textbf{All-tip.} Add both Tip 1 and Tip 2 on the basis of ``None-tip"; for example, when describing Trump's cant, add both ``politician" and ``American" in the political domain to make the prompt more appropriate.

\end{itemize}
\vspace{-2mm}
By observing the effects of these different clues on LLMs, CantCounter can assess the fluctuating changes they induce in recognition and reasoning abilities. This study will help further understand the influence of cues on LLM and provide directions for improving its application and performance.

To answer RQ2, Table \ref{tab2} displays ChatGPT-3.5's accuracy across five domains using different prompt clues. Generally, more clue-related information improves recognition accuracy, as seen in the political domain where All-tip prompts perform significantly better. However, increasing clues doesn't always lead to higher accuracy, possibly due to information redundancy or LLM filter triggering.
Too many clues may reduce accuracy, as seen in the LGBT domain where Tip 1 prompts were less accurate than none-tip prompts.

Our analysis stresses the importance of a balanced clue selection approach to maximize external information usage without compromising accuracy. Thus, choosing appropriate clues in moderate quantities is key to enhancing ChatGPT-3.5's domain-specific performance.

\begin{table*}[t]
\renewcommand{\arraystretch}{1.3}
    \caption{CantCounter stats highlight top performances(\%): Multiple-choice excels in QTM (45.38\%), Zero-shot learning shines in SLM (52.13\%), and All-tip prevails in PCM (29.11\%). ``A" represents ``Abstractive". ``Y/N" represents ``Yes/No". ``Mc" represents ``Multiple-choice". ``Zs" represents ``Zero-shot". ``Os" represents ``One-shot". ``NT" represents ``None-tip". ``T1" represents ``Tip 1". ``T2" represents ``Tip 2". ``AllT" represents ``All-tip".}
\label{tab2}
\centering
\small
\begin{tabular}{c|ccc|cc|cccc}
\toprule
               & \multicolumn{3}{c|}{\cellcolor{QTM}\textbf{QTM}}         & \multicolumn{2}{c|}{\cellcolor{SLM}\textbf{SLM}}     & \multicolumn{4}{c}{\cellcolor{PCM}\textbf{PCM}}      \\ 
\textbf{Domain} & \cellcolor{QTM}\textbf{A} & \cellcolor{QTM}\textbf{Y/N}  & \cellcolor{QTM}\textbf{Mc} & \cellcolor{SLM}\textbf{Zs} & \cellcolor{SLM}\textbf{Os} & \cellcolor{PCM}\textbf{NT} & \cellcolor{PCM}\textbf{T1}   & \cellcolor{PCM}\textbf{T2}   & \cellcolor{PCM}\textbf{AllT} \\ 
\midrule
\textbf{Politics}        & 26.81      & 22.55  & \textbf{50.64 }         & 42.85    & \textbf{57.15 }  & 19.01   & 24.75  & 25.19  & \textbf{31.05 } \\
\textbf{Drugs}           & 21.16      & 22.41  & \textbf{56.43 }         & \textbf{55.41 }   & 44.59   & 17.32   & 27.43  & 25.47  & \textbf{29.78 } \\
\textbf{Racism}          & 29.05      & 27.60  & \textbf{43.35 }         & 41.39    & \textbf{58.61 }  & 11.22   & 19.63  & \textbf{37.50 } & 31.66  \\
\textbf{Weapons}         & \textbf{50.89 }     & 16.20  & 32.91          & \textbf{54.96 }   & 45.04   & 18.73   & \textbf{28.11 } & 25.27  & 27.90  \\
\textbf{LGBT}            & 34.41      & 25.75  & \textbf{39.84 }         & \textbf{59.78 }   & 40.22   & 22.58   & 22.10  & \textbf{28.53 } & 26.79  \\
\textbf{Total}           & 31.71      & \textbf{22.91 }   &  \textcolor{red}{\textbf{45.38 }}         & \textcolor{red}{\textbf{52.13 }}   & 47.87   & 19.03   & 24.61  & 27.24  & \textcolor{red}{\textbf{29.11 }} \\ 
\bottomrule
\end{tabular}\vspace{-2mm}
\end{table*}

\begin{table*}[htbp]
\renewcommand{\arraystretch}{1.3}
    \caption{Zero-shot learning and One-shot learning Q\&A accuracy in CantCounter for GPT-4, Bard, New Bing, and SparkDesk. ``Acc" represents ``Accuracy Rate". ``Rej" represents ``Rejection Rate". ``Don't know" represents ``'I don't know' Rate". }
\label{tab3}
\centering
\footnotesize
\begin{tabular}{c|ccc|ccc}

\toprule
          & \multicolumn{3}{c|}{\textbf{Zero-shot learning}}         & \multicolumn{3}{c}{\textbf{One-shot learning}}          \\ 

          & \textbf{Acc  } & \textbf{Rej} & \textbf{Don't know} & \textbf{Acc  } & \textbf{Rej} & \textbf{Don't know} \\
\midrule
\textbf{ChatGPT-3.5}\cite{TP-OpenAI}  & \textbf{47.61 }       & 4.66          & 39.91       & 45.52        & 1.63          & 46.45       \\
\textbf{GPT-4}\cite{TP-GPT-4}     & 27.27        & \textcolor{red}{\textbf{0.00 }}         & \textcolor{red}{\textbf{70.45 }}      & 50.00        & \textcolor{red}{\textbf{0.00 }}         & 34.09       \\
\textbf{Bard}\cite{TP-Bard}      & 47.73        & 4.55          & 13.64       & 65.91        & \textbf{15.91 }        & 6.82        \\
\textbf{New Bing}\cite{TP-NewBing}  & \textbf{50.00 }       & 11.36         & 34.09       & 50.00        & \textbf{36.36 }        & 2.27        \\
\textbf{SparkDesk}\cite{TP-SparkDesk} & \textbf{29.55 }       & 45.45         & 9.09        & 20.45        & \textbf{68.18 }        & 2.27        \\ 
\bottomrule
\end{tabular}\vspace{-3mm}
\end{table*}
\vspace{-3mm}
\subsection{Comparison with other LLMs}\vspace{-2mm}

In our study, we examine several LLMs alongside ChatGPT-3.5 to address RQ3, including GPT-4\cite{TP-OpenAI}, New Bing \cite{TP-NewBing}, Bard \cite{TP-Bard}, Claude \cite{TP-Claude}, ERNIE \cite{TP-ERNIE}, and SparkDesk \cite{TP-SparkDesk}. While ERNIE is optimized for Chinese content, translating cant prompts may compromise their subtlety and effectiveness. Moreover, ERNIE's frequent account suspensions hindered extensive trials \cite{TP-ERNIEtest}. Claude's sensitive content handling also led to account suspensions \cite{TP-Claude}. Thus, we focus on comparing and validating four other LLMs: GPT-4, Bard, New Bing, and SparkDesk. Table \ref{tab3} presents ratios of correct answers, refused answers, and ``I don't know" responses. Interestingly, GPT-4 consistently responds in all situations, avoiding refusal to answer. This contrasts with other models that often refuse to respond due to content filtering. GPT-4's tendency to use ``I don't know" may stem from our controlled comparisons in the QTM and PCM methods, particularly in Multiple-choice scenarios. Conversely, other LLMs tend to refuse to answer, likely due to content categorization by filters and classifiers. SparkDesk exhibits the highest refusal rate, possibly due to overly strict filters. Furthermore, One-shot learning models are more prone to refusal to answer, as they rely on context understanding, potentially triggering filters. These findings offer insights into the performance of these LLMs across different learning tasks, informing future research directions.

\vspace{-3mm}
\subsection{Takeaways}\vspace{-2mm}
We observe varying accuracy across different Q\&A-Query types (RQ1), with Multiple-choice tasks being most accurate and Yes/No tasks the least. In sensitive domains, Zero-shot learning performs better than One-shot learning (RQ2). Increasing prompt clues improves cant identification accuracy (RQ2). More recent LLM models consistently avoid refusing to answer (RQ3), but they are more likely to refuse answering questions related to racism compared to LGBT (RQ4).

\vspace{-3mm}
\section{Conclusion}\vspace{-2mm}
This paper presents the first comprehensive evaluation of LLM's reasoning capability using cants or dark jargons. We created two domain-specific datasets: Cant and Scene datasets, and developed an evaluation framework to assess LLM's reasoning abilities through cant comprehension. We proposed a four-stage strategy - Fine-Tuning, Co-Tuning, Data-Diffusion, and Data-Analysis - to address cross-matching and complex data calculation problems. Our experiments reveal varying comprehension levels of LLM under different question types (Abstractive, Yes/No, Multiple-choice), sample learning methods (Zero-shot/One-shot learning), and prompt clues (None-tip, Tip1, Tip2, All-tip). Additionally, across different domains (Politics, Drugs, Racism, Weapons, LGBT), different LLMs (GPT-3.5, GPT-4, New Bing, Bard, SparkDesk) demonstrate varying refusal rates to answer questions. Our findings provide insights for the security research community into LLM's reasoning capabilities regarding ``cant", emphasizing the importance of implementing effective safety filters and measures for screening potentially hazardous LLM-generated content.

\bibliographystyle{unsrt}
\bibliography{cant}

\begin{thebibliography}{10}

\bibitem{TP-OpenAI}
OpenAI.
\newblock \url{https://openai.com/chatgpt}.

\bibitem{ray2023chatgpt}
Partha~Pratim Ray.
\newblock Chatgpt: A comprehensive review on background, applications, key challenges, bias, ethics, limitations and future scope.
\newblock {\em Internet of Things and Cyber-Physical Systems}, 2023.

\bibitem{roy2023generating}
Sayak~Saha Roy, Krishna~Vamsi Naragam, and Shirin Nilizadeh.
\newblock Generating phishing attacks using chatgpt.
\newblock {\em arXiv preprint arXiv:2305.05133}, 2023.

\bibitem{weidinger2022taxonomy}
Laura Weidinger, Jonathan Uesato, Maribeth Rauh, Conor Griffin, Po-Sen Huang, John Mellor, Amelia Glaese, Myra Cheng, Borja Balle, Atoosa Kasirzadeh, et~al.
\newblock Taxonomy of risks posed by language models.
\newblock In {\em Proceedings of the 2022 ACM Conference on Fairness, Accountability, and Transparency}, pages 214--229, 2022.

\bibitem{birks2023linking}
Daniel Birks and Joseph Clare.
\newblock Linking artificial intelligence facilitated academic misconduct to existing prevention frameworks.
\newblock {\em International Journal for Educational Integrity}, 19(1):20, 2023.

\bibitem{li2023multi}
Haoran Li, Dadi Guo, Wei Fan, Mingshi Xu, and Yangqiu Song.
\newblock Multi-step jailbreaking privacy attacks on chatgpt.
\newblock {\em arXiv preprint arXiv:2304.05197}, 2023.

\bibitem{gupta2023chatgpt}
Maanak Gupta, CharanKumar Akiri, Kshitiz Aryal, Eli Parker, and Lopamudra Praharaj.
\newblock From chatgpt to threatgpt: Impact of generative ai in cybersecurity and privacy.
\newblock {\em IEEE Access}, 2023.

\bibitem{TP-OpenAI-classifiers}
OpenAI platform.
\newblock \url{https://platform.openai.com/docs/guides/moderation/overview}.

\bibitem{yuan2018reading}
Kan Yuan, Haoran Lu, Xiaojing Liao, and XiaoFeng Wang.
\newblock Reading thieves' cant: automatically identifying and understanding dark jargons from cybercrime marketplaces.
\newblock In {\em 27th USENIX Security Symposium (USENIX Security 18)}, pages 1027--1041, 2018.

\bibitem{tan2023can}
Yiming Tan, Dehai Min, Yu~Li, Wenbo Li, Nan Hu, Yongrui Chen, and Guilin Qi.
\newblock Can chatgpt replace traditional kbqa models? an in-depth analysis of the question answering performance of the gpt llm family.
\newblock In {\em International Semantic Web Conference}, pages 348--367. Springer, 2023.

\bibitem{rozado2023political}
David Rozado.
\newblock The political biases of chatgpt.
\newblock {\em Social Sciences}, 12(3):148, 2023.

\bibitem{TP-Reddit}
Reddit.
\newblock \url{https://www.reddit.com/}.

\bibitem{TP-4chan}
4chan community.
\newblock \url{https://www.4chan.org/}.

\bibitem{sohail2023decoding}
Shahab~Saquib Sohail, Faiza Farhat, Yassine Himeur, Mohammad Nadeem, Dag~{\O}ivind Madsen, Yashbir Singh, Shadi Atalla, and Wathiq Mansoor.
\newblock Decoding chatgpt: A taxonomy of existing research, current challenges, and possible future directions.
\newblock {\em Journal of King Saud University-Computer and Information Sciences}, page 101675, 2023.

\bibitem{RLHF}
Long Ouyang, Jeffrey Wu, Xu~Jiang, Diogo Almeida, Carroll Wainwright, Pamela Mishkin, Chong Zhang, Sandhini Agarwal, Katarina Slama, Alex Ray, et~al.
\newblock Training language models to follow instructions with human feedback.
\newblock {\em Advances in Neural Information Processing Systems}, 35:27730--27744, 2022.

\bibitem{wang2023adversarial}
Jiongxiao Wang, Zichen Liu, Keun~Hee Park, Muhao Chen, and Chaowei Xiao.
\newblock Adversarial demonstration attacks on large language models.
\newblock {\em arXiv preprint arXiv:2305.14950}, 2023.

\bibitem{zhangyang-Why-so-toxic}
Wai~Man Si, Michael Backes, Jeremy Blackburn, Emiliano De~Cristofaro, Gianluca Stringhini, Savvas Zannettou, and Yang Zhang.
\newblock Why so toxic? measuring and triggering toxic behavior in open-domain chatbots.
\newblock In {\em Proceedings of the 2022 ACM SIGSAC Conference on Computer and Communications Security}, pages 2659--2673, 2022.

\bibitem{yuan2023gpt}
Youliang Yuan, Wenxiang Jiao, Wenxuan Wang, Jen-tse Huang, Pinjia He, Shuming Shi, and Zhaopeng Tu.
\newblock Gpt-4 is too smart to be safe: Stealthy chat with llms via cipher.
\newblock {\em arXiv preprint arXiv:2308.06463}, 2023.

\bibitem{ZFXK202103013}
Zhang~Li Feng~Xiaojin.
\newblock Development trend and identification path of drug-related cryptic language under the background of "internet plus".
\newblock {\em Journal of Political Science and Law}, 38(107-118), 2021.

\bibitem{sourdot1991argot}
Marc Sourdot.
\newblock Argot, jargon, jargot.
\newblock {\em Langue fran{\c{c}}aise}, (90):13--27, 1991.

\bibitem{wu2018slangsd}
Liang Wu, Fred Morstatter, and Huan Liu.
\newblock Slangsd: building, expanding and using a sentiment dictionary of slang words for short-text sentiment classification.
\newblock {\em Language Resources and Evaluation}, 52:839--852, 2018.

\bibitem{WHXU201402006}
Qu~Yanbin.
\newblock Grammar summary of chinese folk secret language (lingo) (part 1).
\newblock {\em Cultural Journal}, (26-33), 2014.

\bibitem{yan2017building}
Zhao Yan, Nan Duan, Peng Chen, Ming Zhou, Jianshe Zhou, and Zhoujun Li.
\newblock Building task-oriented dialogue systems for online shopping.
\newblock In {\em Proceedings of the AAAI Conference on Artificial Intelligence}, volume~31, 2017.

\bibitem{dasigi2021dataset}
Pradeep Dasigi, Kyle Lo, Iz~Beltagy, Arman Cohan, Noah~A Smith, and Matt Gardner.
\newblock A dataset of information-seeking questions and answers anchored in research papers.
\newblock {\em arXiv preprint arXiv:2105.03011}, 2021.

\bibitem{rogers2023qa}
Anna Rogers, Matt Gardner, and Isabelle Augenstein.
\newblock Qa dataset explosion: A taxonomy of nlp resources for question answering and reading comprehension.
\newblock {\em ACM Computing Surveys}, 55(10):1--45, 2023.

\bibitem{TP-Eric-Patterson}
X~Corp.
\newblock \url{https://drugabuse.com/addiction/list-street-names-drugs/}.

\bibitem{TP-X-Corp}
X.
\newblock \url{https://twitter.com/}.

\bibitem{TP-Kaggle}
Kaggle.
\newblock \url{https://www.kaggle.com}.

\bibitem{TP-Hugging-Face}
Hugging Face.
\newblock \url{https://huggingface.co/}.

\bibitem{zhao2020cyber}
Jun Zhao, Qiben Yan, Xudong Liu, Bo~Li, and Guangsheng Zuo.
\newblock Cyber threat intelligence modeling based on heterogeneous graph convolutional network.
\newblock In {\em 23rd international symposium on research in attacks, intrusions and defenses (RAID 2020)}, pages 241--256, 2020.

\bibitem{TP-politics1}
EverybodyWiki Bios~\& Wiki.
\newblock \url{https://en.everybodywiki.com/List_of_nicknames_of_Donald_Trump}.

\bibitem{TP-drug1}
Defining Wellness.
\newblock \url{https://definingwellness.com/resources/drug-slang-word-glossary/}.

\bibitem{TP-Weapon2}
A~Gun Lingo~Glossary for Those Unfamiliar With~Firearms.
\newblock \url{https://lifehacker.com/a-gun-lingo-glossary-for-those-unfamiliar-with-firearms-1825427596}.

\bibitem{TP-race1}
The Racial~Slur Database.
\newblock \url{http://www.rsdb.org/races}.

\bibitem{TP-LGBT1}
Wikipedia.
\newblock \url{https://en.wikipedia.org/wiki/LGBT_slang}.

\bibitem{vaswani2017attention}
Ashish Vaswani, Noam Shazeer, Niki Parmar, Jakob Uszkoreit, Llion Jones, Aidan~N Gomez, {\L}ukasz Kaiser, and Illia Polosukhin.
\newblock Attention is all you need.
\newblock {\em Advances in neural information processing systems}, 30, 2017.

\bibitem{zhong2023study}
Li~Zhong and Zilong Wang.
\newblock A study on robustness and reliability of large language model code generation.
\newblock {\em arXiv preprint arXiv:2308.10335}, 2023.

\bibitem{TP-GPT-4}
GPT-4.
\newblock \url{https://openai.com/research/gpt-4}.

\bibitem{TP-Bard}
Bard-Google.
\newblock \url{https://bard.google.com/}.

\bibitem{TP-NewBing}
NewBing.
\newblock \url{https://www.bing.com/new}.

\bibitem{TP-SparkDesk}
SparkDesk Xunfei-Xinghuo.
\newblock \url{https://xinghuo.xfyun.cn/}.

\bibitem{TP-Claude}
Google.
\newblock \url{https://claude.ai/}.

\bibitem{TP-ERNIE}
ERNIE.
\newblock \url{https://yiyan.baidu.com/welcome}.

\bibitem{TP-ERNIEtest}
ERNIE~Protection Rule.
\newblock https://wanhua.baidu.com/talk/protectionrule.

\end{thebibliography}

\end{document}